\documentclass[11pt,a4paper]{article}
\usepackage[hyperref]{emnlp2020}
\usepackage{times}
\usepackage{latexsym}

\usepackage{microtype}

\aclfinalcopy 
\def\aclpaperid{***} 

\usepackage{graphicx}
\usepackage{subfigure}
\usepackage{times}
\usepackage{amssymb}
\usepackage{amsmath}

\title{Edge-Enhanced Graph Convolution Networks for \\
	Event Detection with Syntactic Relation}

\author{Shiyao Cui$^{1,2}$ \ Bowen Yu$^{1,2}$ \ Tingwen Liu$^{1,2}$\thanks{\hspace{0.15cm}Corresponding   Author} \ Zhenyu Zhang$^{1,2}$ \ Xuebin Wang$^{1,2}$ \and Jinqiao Shi$^{3,1}$  \\
     $^1$Institute of Information Engineering, Chinese Academy of Sciences. Beijing, China \\
     $^2$School of Cyber Security, University of Chinese Academy of Sciences. Beijing, China \\
     $^3$Beijing University of Posts and Telecommunications. Beijing, China \\
     {\tt \{cuishiyao, yubowen, zhangzhenyu1996\}@iie.ac.cn} \\
     {\tt \{liutingwen, wangxuebin\}@iie.ac.cn} \\
     	{\tt shijinqiao@bupt.edu.cn}
            }

\date{}

\begin{document}
\maketitle
\begin{abstract}
  Event detection (ED), a key subtask of information extraction, aims to recognize instances of specific event types in text.
Previous studies on the task have verified the effectiveness of integrating syntactic dependency into graph convolutional networks.
However, these methods usually ignore dependency label information, which conveys rich and useful linguistic knowledge for ED.
In this paper, we propose a novel architecture named Edge-Enhanced Graph Convolution Networks (EE-GCN), which simultaneously exploits syntactic structure and typed dependency label information to perform ED.
Specifically, an edge-aware node update module is designed to generate expressive word representations by aggregating syntactically-connected words through specific dependency types. 
%
Furthermore, to fully explore clues hidden in dependency edges, a node-aware edge update module is introduced, which refines the relation representations with contextual information.
%
These two modules are complementary to each other and work in a mutual promotion way. 
We conduct experiments on the widely used ACE2005 dataset and the results show significant improvement over competitive baseline methods\footnote{Source code of this paper could be obtained from https://github.com/cuishiyao96/eegcned}.
\end{abstract}

\section{Introduction}

\begin{figure}[t!]
	
	\centering
	\includegraphics[width=0.98\columnwidth]{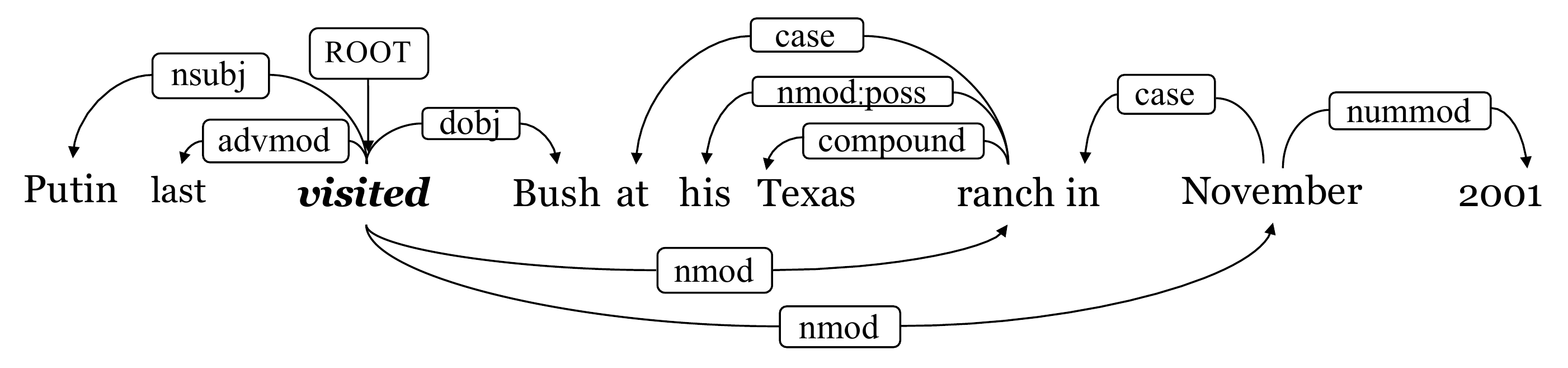}
	\caption{An example of syntactic dependency parsing, which contains an event of \emph{Meet} triggered by ``\emph{visited}''.}\label{fig:example}
\end{figure}

Event Detection (ED) is an important information extraction task that seeks to recognize events of specific types from given text.
Specifically, each event in a sentence is marked by a word or phrase called ``event trigger''.
The task of ED is to detect event triggers  and classify them into specific types of interest. 
Taking Figure~\ref{fig:example} as an example, ED is supposed to recognize the event trigger ``\emph{visited'}' and classify it to the event type \emph{Meet}.

Dependency trees convey rich structural information that is proven useful for ED~\cite{nguyen2018graph,liu-etal-2018-jointly,yan-etal-2019-event}.
Recent works on ED focus on building Graph Convolutional Networks (GCNs) over the dependency tree of a sentence to exploit syntactic dependencies~\cite{nguyen2018graph,liu-etal-2018-jointly,yan-etal-2019-event}.
Compared to sequence-based models, GCN-based models are able to capture non-local syntactic relations that are obscure from the surface form alone~\cite{guo-etal-2019-attention}, and usually achieve better performance.

Nevertheless, existing GCN-based ED methods do not consider dependency labels, which may serve as significant indicators to reveal whether a word is a trigger or not.
As shown in Figure~\ref{fig:example}, the dependency ``nsubj'' (nominal subject) and ``dobj''(direct object) show that ``\emph{Putin}'' and ``\emph{Bush}'' are the subject and object of ``\emph{visited}'' respectively, and the words connected to ``\emph{visited}'' with ``nmod''(noun compound modifier) dependency express when and where the event happened. 
Apparently, such dependency labels constitute an effective evidence to predict the event type of ``\emph{visited}'' as \emph{Meet}.
In addition, our statistical results on the benchmark ACE2005 dataset show that  ``nsubj'', ``dobj'' and ``nmod'' take up 32.2\% of trigger-related dependency labels (2.5\% for each relation on average among all 40 dependency relations), which means that simultaneously modeling syntactic structure and dependency labels can be crucial to make full use of the dependency trees to further improve the performance of ED.

Besides, we also observe that the same dependency label under different context may convey different signals for ED.
Again, taking Figure~\ref{fig:example} as an example: the dependency ``nmod'' connected with ``\emph{ranch}'' indicates \textbf{where} the event happens but another dependency ``nmod'' connected with ``\emph{November}'' points out \textbf{when} the event happens.
Such an observation demonstrates that assigning a single context-independent representation for each dependency label is not enough to express the complex relations between words.
This is to say, the representations of dependency relations should be context-dependent and dynamic, calculated and updated according to a sentential context using a network structure.

To model the above ideas, in this paper, we propose a novel neural architecture named Edge-Enhanced Graph Convolutional Networks (EE-GCN), which explicitly takes advantage of the typed dependency labels with dynamic representations. 
In particular, EE-GCN transforms a sentence to a graph by treating words and dependency labels as nodes and typed edges, respectively.
Accordingly, an adjacency tensor is constructed to represent the graph, where syntactic structure and typed dependency labels are both captured.
To encode the heterogeneous information from the adjacency tensor, EE-GCN simultaneously performs two kinds of propagation learning. 
For each layer, an edge-aware node update module is firstly performed for aggregating information from neighbors of each node through specific edges. 
Then a node-aware edge update module is used to dynamically refine the edge representation with its connected node representations, making the edge representation more informative.
These two modules work in a mutual promotion way by updating each other iteratively.

Our contributions are summarized as follows:
\begin{itemize}
	\item We propose the novel EE-GCN that simultaneously integrate syntactic structure and typed dependency labels to improve neural event detection, and learns to update the relation representations in a context-dependent manner. To the best of our knowledge, there is no similar work in ED.
	
	
	
	\item Experiments conducted on the ACE2005\footnote{https://catalog.ldc.upenn.edu/LDC2006T06} benchmark show that EE-GCN achieves SOTA performance. Further analysis confirms the effectiveness and efficiency of our model.

\end{itemize}

%

\section{Related Works}

In earlier ED studies, researchers focused on leveraging various kinds of linguistic features and manually designed feature for the task. However, all the feature-based methods depend on the quality of designed features from a pre-processing step.

Most recent works have focused on leveraging neural networks in this task~\cite{chen-etal-2015-event,nguyen-grishman-2015-event,nguyen-etal-2016-joint-event,ghaeini-etal-2016-event,feng-etal-2016-language}. The existing approaches can be categorized into two classes:
The first class is to improve ED through special learning techniques including adversarial training~\cite{hong-etal-2018-self}, knowledge distillation~\cite{liu2019exploiting,lu-etal-2019-distilling} and model pre-training~\cite{yang-etal-2019-exploring}.
The second class is to improve ED by introducing extra resource, such as argument information~\cite{liu-etal-2017-exploiting}, document information~\cite{duan-etal-2017-exploiting,zhao-etal-2018-document,chen-etal-2018-collective}, multi-lingual information~\cite{liu2018event,liu2019exploiting}, knowledge base~\cite{liu-etal-2016-leveraging,chen-etal-2017-automatically} and syntactic information~\cite{sha2018jointly}.

Syntactic information plays an important role in ED.
~\citeauthor{sha2018jointly}~\shortcite{sha2018jointly} exploited a dependency-bridge recurrent neural network to integrate the dependency tree into model.
~\citeauthor{orr-etal-2018-event}~\shortcite{orr-etal-2018-event} proposed a directed-acyclic-graph GRU model to introduce syntactic structure into sequence structure.
With the rise of GCN~\cite{kipf2017semi}, researchers proposed to transform the syntactic dependency tree into a graph and employ GCN to conduct ED through information propagation over the graph~\cite{nguyen2018graph,liu-etal-2018-jointly,yan-etal-2019-event}.
%
%
Although these works use syntax structures, few of them take dependency label information into consideration, which we, here, demonstrate its importance.
How to effectively leverage the typed dependency information still remains a challenge in this task.


\begin{figure*}[t!]
	\centering
	\includegraphics[width=0.95\linewidth]{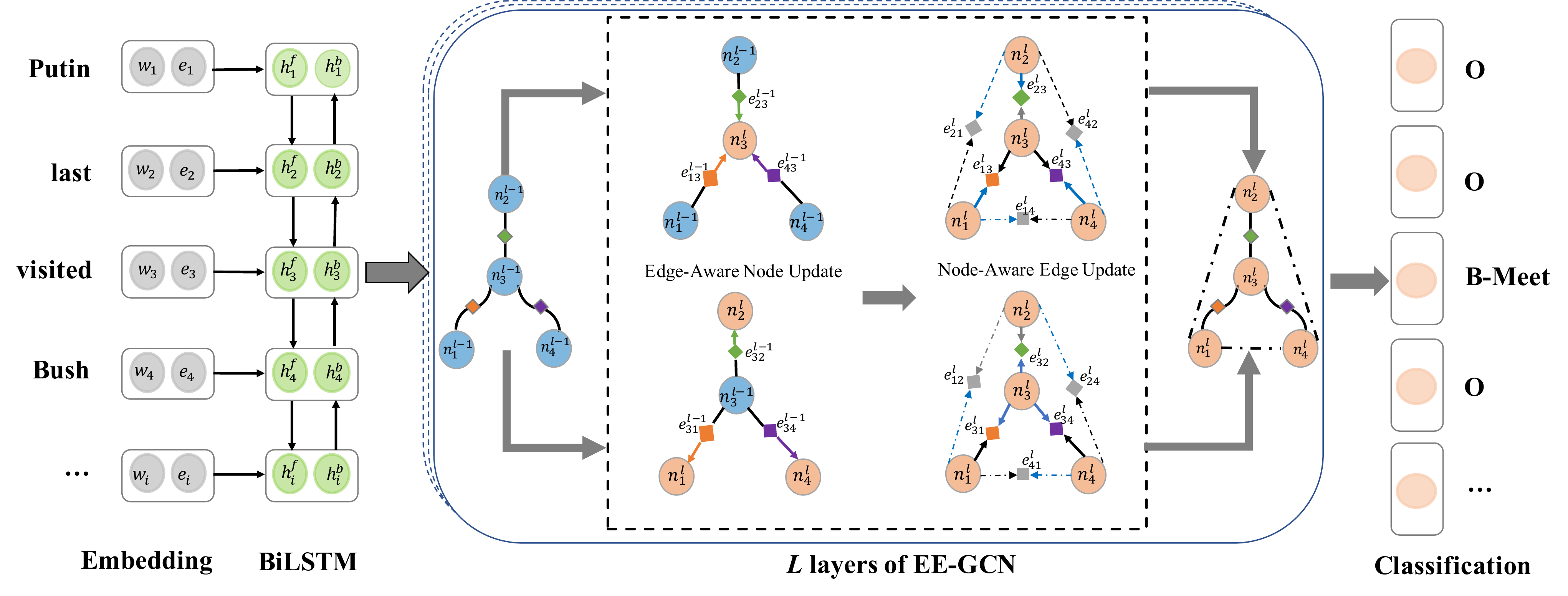}
	\caption{Illustration of EE-GCN event detection architecture. After embedding and BiLSTM layer, $L$ layers of EE-GCN are stacked to learn word representation for sequence labeling. EE-GCN contains two modules: Edge-Aware Node Update Module first aggregates information from neighbors of each node through specific edge, and Node-Aware Edge Update module refines the edge representation with its connected nodes.
	}\label{fig:modelframe}
\end{figure*}

\section{Problem Statement}

In this section, we formally describe the event detection problem.
Following previous works~\cite{chen-etal-2015-event,nguyen-etal-2016-joint-event,liu-etal-2017-exploiting,chen-etal-2018-collective,yan-etal-2019-event}, we formulate event detection as a sequence labeling task. 
Each word is assigned a label that contributes to event annotation. 
Tag ``O'' represents the ``Other'' tag, which means that the corresponding word is irrelevant of the target events. 
In addition to ``O'', the other tags consist of two parts: the word position in the trigger and the event type. 
We use the ``BI'' (Begin, Inside) signs to represent the position information of a word in the event trigger.
The event type information is obtained from a pre-defined set of events.
Thus, the total number of tags is $2 \times N_{EventType} + 1$, where $N_{EventType}$ is the number of predefined event types.

\section{Methods}

Figure~\ref{fig:modelframe} gives an illustration of EE-GCN based event detection architecture, which is mainly composed of three components: the Input Layer, the Edge-Enhanced GCN layer and the Classification Layer.
Next, we detail all components sequentially from bottom to top.

\subsection{Input Layer}

Let $S=\{w_1, w_2, ..., w_n\}$ denote an $n$-word sentence, we first transform each word to a real-valued vector $\boldsymbol{\rm{x}}_i$ by concatenating the following vectors:
\begin{itemize}
	
	\item Word embedding $\boldsymbol{\rm{w}}_i$: it captures the meaningful semantic regularity of word. Following previous works~\cite{chen-etal-2018-collective,yan-etal-2019-event}, we use the word embedding pre-trained by Skip-gram on the NYT Corpus.
	\item Entity type embedding $\boldsymbol{\rm{e}}_i$: entities in the sentence are annotated with BIO schema and we map each entity type label to a real-valued embedding by looking up an embedding table.
\end{itemize}

Thus, the input embedding of $w_i$ can be defined as  $\boldsymbol{\rm{x}}_i = [\boldsymbol{\rm{w}}_i; \boldsymbol{\rm{e}}_i]\in \mathbb{R}^{d_w+d_e}$ , where $d_w$ and $d_e$ denote the dimension of word embedding and entity type embedding respectively.
Then, a BiLSTM layer is adopted to capture the contextual information for each word. 
For simplicity, we denote the contextualized word representations as $\mathbf{S}=[\mathbf{h}_1, \cdots,\mathbf{h}_n]$, where $\mathbf{S} \in \mathbb{R}^{n \times d}$ are used as initial node features in EE-GCN.

\subsection{Edge-Enhanced Graph Convolutional Networks} 

In this subsection, we start by introducing the baseline GCN model, and then present the proposed EE-GCN, which can make full use of dependency label features for better  representation learning.

\subsubsection{Vanilla Graph Convolutional Network}

GCN~\cite{kipf2017semi}, which is capable of encoding graphs, is an extension of convolutional neural network. 
For an $L$-layer GCN where $l\in[1,\cdots,L]$, if we denote $\mathbf{H}^{l-1}$ the input state and $\mathbf{H}^{l}$ the output state of the $l$-th layer, the graph convolutional operation can be formulated as:
\begin{equation}
\begin{split}
\boldsymbol{\rm{H}}^{l} &=  \text{GCN}(\boldsymbol{\rm{A}} \text{, } \boldsymbol{\rm{H}}^{l-1} \text{, } \boldsymbol{\rm{W}}) \\
&=  \sigma(\boldsymbol{\rm{A}} \boldsymbol{\rm{H}}^{l-1} \boldsymbol{\rm{W}}),
\end{split}
\end{equation}

\noindent where $\boldsymbol{\rm{A}} \in \mathbb{R}^{n \times n}$ is an adjacency matrix expressing connectivity between nodes, $\boldsymbol{\rm{W}}$ is a learnable convolutional filter and $\sigma$ denotes a nonlinear activation function, e.g., ReLU.


Previous GCN-based ED methods~\cite{nguyen2018graph,liu-etal-2018-jointly,yan-etal-2019-event} transform dependency tree to a graph according to syntactic connectivity, with each word in the sentence regarded as a node.
The graph is represented by an $n \times n$ adjacency matrix $\mathbf{A}$ through enumerating the graph, where $\boldsymbol{\rm{A}}_{ij} = 1$ if there is a syntactic dependency edge between node $i$ and node $j$,  otherwise $\boldsymbol{\rm{A}}_{ij} = 0$.
Obviously, such approaches use a binary adjacent matrix as structural information, and omit typed dependency label features, which can be potentially useful for ED as discussed in the introduction.
%
%
It is supposed to be mentioned that why these methods ignore typed dependency labels.
An intuitive way for vanilla GCN to exploit these labels is to encode different types of dependency relation with different convolutional filters, which is similar to RGCN~\cite{kipf2017semi}.
However, RGCN suffers from over-parameterization, where the number of parameters grows rapidly with the number of relations.
Given that there exists approximately 40 types of dependency relations and the size of ED dataset is just moderate, models with large amount of parameters are likely to overfit, for which previous works for ED ignore typed dependency labels.

\subsubsection{Edge-Enhanced GCN}

Edge-Enhanced GCN (EE-GCN) is an extension of the vanilla GCN mentioned above, which incorporates typed dependency label information into the feature aggregation process to obtain better representations.
Specifically, EE-GCN constructs an adjacency tensor $\mathbf{E} \in \mathbb{R}^{n\times n \times p}$  to describe the graph structure instead of the binary adjacency matrix used in the vanilla GCN, where $\mathbf{E}_{i,j,:}\in \mathbb{R}^{p}$ is the $p$-dimensional relation representation between node $i$ and node $j$, and $p$ can also be understood as the number of channels in the adjacency tensor.
%
%
%
Formally, $\mathbf{E}$ is initialized according to the dependency tree, if a dependency edge exists between $w_i$ and $w_j$ and the dependency label is $r$,  then $\mathbf{E}_{i,j,:}$ is initialized to the embedding of $r$ obtained from a trainable embedding lookup table, otherwise we initialize $\mathbf{E}_{i,j,:}$ with a $p$-dimensional all-zero vector.
%
%
%
Following previous works ~\cite{marcheggiani-titov-2017-encoding,zhang-etal-2018-graph,guo-etal-2019-attention}, we initialize $\mathbf{E}$ based on an undirectional graph, which means that $\mathbf{E}_{i,j,:}$ and $\mathbf{E}_{j,i,:}$ are initialized as the same embedding.
For the ROOT node in the dependency tree, we add a self loop to itself with a special relation ``ROOT''. 

In order to fully leverage the adjacency tensor and effectively mine latent relation information beyond the dependency labels, two modules are implemented at each layer $l$ of EE-GCN to update the node representations ($\mathbf{H}$) and edge representations ($\mathbf{E}$) mutually through information aggregation:
\begin{equation}
\boldsymbol{\rm{H}}^l,\boldsymbol{\rm{E}}^l = \text{EE-GCN}(\boldsymbol{\rm{E}}^{l-1},\boldsymbol{\rm{H}}^{l-1}).
\label{con:egcn}
\end{equation}

\noindent\textbf{Edge-Aware Node Update Module}

With words in sentence interpreted as nodes in graph, edge-aware node update (EANU) module updates the representation for each node by aggregating the information from its neighbors through the adjacency tensor.
Mathematically, this operation can be defined as follows:
%
%
%
%
\begin{equation}
\begin{split}
\boldsymbol{\rm{H}}^{l} &= \text{EANU}(\boldsymbol{\rm{E}}^{l-1} , \boldsymbol{\rm{H}}^{l-1}) \\
&= \sigma(\text{Pool}(\boldsymbol{\rm{H}}^{l}_1, \boldsymbol{\rm{H}}^{l}_2, ..., \boldsymbol{\rm{H}}^{l}_p)).\\
\end{split}
\end{equation}
Specifically, the aggregation is conducted channel by channel in the adjacency tensor as follows:
\begin{equation}
\begin{split}
\boldsymbol{\rm{H}}^{l}_i=\boldsymbol{\rm{E}}^{l-1}_{:,:,i}\boldsymbol{\rm{H}}^{l-1}\boldsymbol{\rm{W}},
\end{split}
\end{equation}	

\noindent where $\mathbf{E}^{l-1} \in \mathbb{R}^{n\times n \times p}$ is the adjacency tensor from initialization or last EE-GCN layer, $\mathbf{E}^{l-1}_{:,:,i} \in \mathbb{R}^{n\times n}$ denotes the ${i}_{th}$ channel slice of $\mathbf{E}^{l-1}$, $\mathbf{H}^{0}$ is the output of BiLSTM, $\mathbf{W} \in \mathbb{R}^{d\times d}$ is a learnable filter, $d$ is the dimension of node representation, and $\sigma$ is the ReLU activation function.
%
%
%
A mean-pooling operation is applied to compress features since it covers information from all channels.

\noindent\textbf{Node-Aware Edge Update Module}

In the original adjacency tensor, the relation representation between words is initialized to the dependency label embedding.
However, as mentioned in the introduction, the same dependency label under different context may convey different signals for ED, thus assigning a single context-independent representation for each dependency label is not enough to express the complex relations between words.
To address this issue, we propose a novel node-aware edge update (NAEU) module to dynamically calculate and update edge representations according to the node context.
Formally, the NAEU operation is defined as:
\begin{equation}
\begin{split}
\boldsymbol{\rm{E}}^{l}_{i,j,:} &= \text{NAEU}(\boldsymbol{\rm{E}}^{l-1}_{i,j,:}, \boldsymbol{\rm{h}}^{l}_{i}, \boldsymbol{\rm{h}}^{l}_{j})  \\
&= \boldsymbol{\rm{W}}_u[\boldsymbol{\rm{E}}^{l-1}_{i,j,:}\oplus 
\boldsymbol{\rm{h}}^{l}_{i}\oplus\boldsymbol{\rm{h}}^{l}_{j}], i,j \in [1,n],
\end{split}
\end{equation}

\noindent where $\oplus$ means the concatenation operator, $\mathbf{h}^{l}_{i}$ and $\mathbf{h}^{l}_{j}$ denote the representations of node $i$ and node $j$ in the $l_{th}$ layer after EANU operation, respectively, $\mathbf{E}^{l-1}_{i,j,:} \in \mathbb{R}^{p}$ is the relation representation between node $i$ and node $j$, $\mathbf{W}_u \in \mathbb{R}^{(2\times d + p)\times p}$ is a learnable transformation matrix.
This operation refines the adjacency tensor in a context-dependent manner, so that the latent relation information expressed in the node representations can be effectively mined and injected to the adjacency tensor.
And the adjacency tensor is no longer constrained to just convey the dependency label information, obtaining more representation power.
%
The updated adjacency tensor is fed into the next EE-GCN layer to perform another round of edge-aware node update, and such mutual update process can be be stacked over $L$ layers.


\subsection{Classification Layer}

After aggregating word (node) representations from each layer of EE-GCN, we finally feed the representation of each word into a fully-connected network, which is followed by a softmax function to compute distribution $p(t|\boldsymbol{\rm{h}})$ over all event types:
\begin{equation}
p(t|\boldsymbol{\rm{h}}) = {\rm softmax} (\boldsymbol{\rm{W}}_t \boldsymbol{\rm{h}} + \boldsymbol{\rm{b}}_t ),
\end{equation}
\noindent where $\mathbf{W}_t$ maps the word representation $\mathbf{h}$ to the feature score for each event type and $\boldsymbol{\rm{b}}_t$ is a bias term.
After softmax, event label with the largest probability is chosen as the classification result.

\subsection{Bias Loss Function}

Following popular choices~\cite{chen-etal-2018-collective,yan-etal-2019-event}, we adopt a bias loss function to strengthen the influence of event type labels during training, since the number of ``O"  tags is much lager than that of event type tags.
The bias loss function is formulated as follows:
\begin{equation}
\begin{split}
J(\theta ) =& - \sum_{i=1}^{N_s}\sum_{j=1}^{n_i}\log{p(y_{j}^{t}|s_{i},\theta)} \cdot {I(O)}  \\
& +{\alpha}\log{p(y_{j}^{t}|s_{i},\theta)} \cdot {(1-I(O))},
\end{split}
\end{equation}
\noindent where ${N_s}$ is the number of sentences, ${n_i}$ is the number of words in the $i_{th}$ sentence; 
$I(O)$ is a switching function to distinguish the loss of tag ``O" and event type tags. 
It is defined as follows:
\begin{equation}
I(O) = 
\begin{cases}
1, & \mbox{if tag is  ``O"} \\
0, & \mbox{otherwise}
\end{cases}	,
\label{eq.rw}
\end{equation}
where $\alpha$ is the bias weight. The larger the $\alpha$ is, the greater the influence of event type tags on the model.

\section{Experiments}

\subsection{Dataset and Evaluation Metrics}

We conduct experiments on the ACE2005 dataset, which is the standard supervised dataset for event detection.
The Stanford CoreNLP toolkit\footnote{http://nlp.stanford.edu/software/stanford-english-corenlp-2018-10-05-models.jar} is used for dependency parsing.
ACE2005 contains 599 documents annotated with 33 event types.
We use the same data split as previous works~\cite{chen-etal-2015-event,nguyen-etal-2016-joint-event,liu-etal-2017-exploiting,chen-etal-2018-collective,yan-etal-2019-event} for train, dev and test set, and describe the details in the supplementary material (Data.zip).
We evaluate the models using the official scorer in terms of the Precision (P), Recall (R) and $F_1$-score\footnote{https://github.com/yubochen/NBTNGMA4ED/}.



\subsection{Hyper-parameter Setting}

The hyper-parameters are manually tuned on the dev set. 
We adopt word embeddings pre-trained on the NYT corpus with the Skip-gram algorithm and the dimension is 100. 
The entity type and dependency label embeddings are randomly initialized.
We randomly initialize the entity type and dependency label embeddings with 25- and 50- dimension vectors.
The hidden state size of BiLSTM and EE-GCN are set to 100 and 150, respectively. 
Parameter optimization is performed using SGD with learning rate 0.1 and batch size 30.
We use L2 regularization with a parameter of 1e-5 to avoid overfitting.
Dropout is applied to word embeddings and hidden states with a rate of 0.6.
The bias parameter $\alpha$ is set to 5.
The max length of sentence is set to be 50 by padding shorter sentences and cutting longer ones. 
The number of EE-GCN layers is 2, which is the best-performing depth in pilot studies.
We ran all the experiments using Pytorch 1.1.0 on Nvidia Tesla P100 GPU, with Intel Xeon E5-2620 CPU.



\subsection{Baselines}
In order to comprehensively evaluate our proposed EE-GCN model, we compare it with a range of baselines and state-of-the-art models, which can be categorized into three classes: feature-based, sequence-based and GCN-based.

	
	\textbf{Feature-based models} use human designed features to perform event detection. 1) \textbf{MaxEnt} is proposed by~\newcite{li-etal-2013-joint} using lexical and syntactic features; 2) \textbf{CrossEntity} is proposed by \newcite{hong-etal-2011-using} using cross-entity information to detect events.
	
	\textbf{Sequence-based models} operate on the word sequences. 1) \textbf{DMCNN}\cite{chen-etal-2015-event} builds a dynamic multi-pooling convolutional model to learn sentence features; 2) \textbf{JRNN}~\cite{nguyen-etal-2016-joint-event} employs bidirectional RNN for the task; 3) \textbf{ANN-AugAtt}~\cite{liu-etal-2017-exploiting} uses annotated event argument information with supervised attention, where words describing Time, Place and Person of events get larger attention score; 4) \textbf{dbRNN}~\cite{sha2018jointly} adds dependency arcs with weight to BiLSTM to make use of tree structure and sequence structure simultaneously; 5) \textbf{HBTNGMA}~\cite{chen-etal-2018-collective} applies hierarchical and bias tagging networks to detect multiple events in one sentence collectively.
	
	\textbf{GCN-based models} build a Graph Convolutional Network over the dependency tree of a sentence to exploit syntactical information. 1) \textbf{GCN-ED}~\cite{nguyen2018graph} is the first attempt to explore how to effectively use GCN in event detection;2) \textbf{JMEE}~\cite{liu-etal-2018-jointly} enhances GCN with self-attention and highway network to improve the performance of GCN for event detection; 3) \textbf{RGCN}~\cite{schlichtkrull2018modeling}, which models relational data with relation-specific adjacency matrix and convolutional filter, is originally proposed for knowledge graph completion. We adapt it to the task of event detection by using the same classification layer and bias loss with our model; 4) \textbf{MOGANED}~\cite{yan-etal-2019-event} improves GCN with aggregated attention to combine multi-order word representation from different GCN layers, which is the state-of-the-art method on the ACE2005 dataset.
	

 \subsection{Overall Performance}
\label{sec:performance}

\begin{table}
	\small
	\centering
	\begin{tabular}{lccc}
		\hline \textbf{Model} & \textbf{P} & \textbf{R} & \textbf{$F_1$} \\ \hline
		MaxEnt~\cite{li-etal-2013-joint}       & 74.5  & 59.1  & 65.9   \\
		CrossEntity~\cite{hong-etal-2011-using}       & 72.9  & 64.3  & 68.3   \\
		\hline
		DMCNN~\cite{chen-etal-2015-event}       & 75.6  & 63.6 & 69.1   \\
		JRNN~\cite{nguyen-etal-2016-joint-event}       &66.0  & 73.0 & 69.3    \\
		ANN-AugAtt~\cite{liu-etal-2017-exploiting}     & 78.0  & 66.3 & 71.7   \\
		dbRNN$^{\dag}$~\cite{sha2018jointly}    & 74.1  & 69.8 & 71.9     \\
		HBTNGMA~\cite{chen-etal-2018-collective}       & 77.9  & 69.1 & 73.3      \\
		\hline
		GCN-ED$^{\dag}$    & 77.9  & 68.8 & 73.1    \\
		JMEE$^{\dag}$~\cite{liu-etal-2018-jointly}    & 76.3  & 71.3 & 73.7     \\	
		RGCN$^{\dag \ddag}$~\cite{schlichtkrull2018modeling}   & 68.4  & \textbf{79.3} & 73.4    \\
		MOGANED$^{\dag}$~\cite{yan-etal-2019-event}    & \textbf{79.5}  & 72.3 & 75.7     \\
		\hline
		EE-GCN$^{\dag \ddag}$  &76.7  & 78.6  & \textbf{77.6}   \\
		
		\hline
	\end{tabular}
	\caption{\label{tab:performance} Results on ACE2005. ${\dag}$ means model only using dependency structure and ${\dag \ddag}$ denotes model using syntactic dependency structure and typed dependency label simultaneously; Bold marks the highest score among all models. Moreover, the Wilcoxon’s test shows that significant difference ($p < 0.05$) exists between our model and the previous SOTA MOGANED.}
\end{table}

We report our experimental results on the ACE2005 dataset in Table~\ref{tab:performance}. 
It is shown that our model, EE-GCN, outperforms all the baselines and achieves state-of-the-art $F_1$-score.
We attribute the performance gain to two aspects:
1) The introduction of typed dependency label. 
EE-GCN outperforms all existing GCN-based models which only utilize syntactic structure and ignore the specific typed dependency labels, this demonstrates that the type of dependency label is capable of providing key information for event detection.
2) The design of context-dependent relation representation. 
Compared with the baseline RGCN which also exploits both syntactic structure and dependency labels, EE-GCN still improves by an absolute margin of 4.2\%. 
We consider that it is because RGCN distinguishes different dependency labels with different convolution filters, thus the same dependency label maintains the same representation regardless of the different context. 
As a result, the potential relation information expressed beyond dependency labels is not fully exploited. 
By contrast, our EE-GCN model learns a context-dependent relation representation during information aggregation process with the help of the node-aware edge update module, and thus better captures the information under relations between words. 
%

We also observe that EE-GCN gains its improvements mainly on Recall, and we hypothesize that this is because EE-GCN introduces dependency label, which help to capture more fine-grained trigger-related features, thus more triggers would be detected.
Meanwhile, MOGANED surpasses EE-GCN on Precision, which could be explained as the original paper analyzed that since MOGANED exploited GAT(GCN with attention) as basic encoder, the attention mechanism helps to predict event triggers more precisely.

Additionally, we notice that EE-GCN performs remarkably better than all sequence-based neural models that do not use dependency structure, which clearly demonstrates that the reasonable use of syntactic dependency information can indeed improve the performance of event detection.
When comparing EE-GCN with dbRNN which adds weighted syntactic dependency arcs to BiLSTM, our model gains improvement on both P and R. 
This phenomenon illustrates that GCN is capable of modeling dependency structure more effectively and the multi-dimensional embedding of dependency label in EE-GCN learn more information than just a weight in dbRNN.

\section{Analysis}

\subsection{Ablation Study}

To demonstrate the effectiveness of each component, we conduct an ablation study on the ACE2005 dev set as Table~\ref{tab:ablation} shows \footnote{Note that the $F_1$ score of model on the ACE2005 dev set is significantly lower than that on the test set. We guess the performance difference comes from the domain gap that the ACE2005 dev set and test set are collected from different domains~\cite{nguyen-grishman-2015-event}.}
%
1) -- Typed Dependency Label (TDL): to study whether the typed dependency labels contribute to the performance improvement, we initialize each $\mathbf{E}_{i,j,:}$ in the adjacency tensor $\mathbf{E}$ as the same vector if there is a syntactic dependency edge between node $i$ and node $j$, thus the typed dependency label information is removed. As a result, the $F_1$-score drops by 0.5\% absolutely, which demonstrates that typed dependency label information plays an important role in EE-GCN.
2) -- Node-Aware Edge Update Module (NAEU): removing node-aware edge update module hurts the result by 0.99\% $F_1$-score, which verifies that the context-dependent relation representations provide more evident information for event detection than the context-independent ones.
3) -- TDL \& NAEU: we remove edge-aware node update module and node-aware edge update module simultaneously, then the model is degenerated to the vanilla GCN. We observe that the performance reduces by 1.69\%, which again confirms the effectiveness of our model.
4) -- Multi-dimensional Edge representation (MDER): when we set the dimension of relation representation to 1, this is to compress the adjacency tensor $\boldsymbol{\rm{E}} \in \mathbb{R}^{n\times n \times p}$ to be $\boldsymbol{\rm{E}} \in \mathbb{R}^{n \times n \times 1}$, the $F_1$-score drops by 0.77 \% absolutely, which indicates that the multi-dimensional representation is more powerful to capture information than just a scalar parameter or weight.
%
5) --BiLSTM: BiLSTM is removed before EE-GCN and the performance drops terribly. This illustrates that BiLSTM capture important sequential information which GCN misses. Therefore, GCN and BiLSTM are complementary to each other for event detection.

\begin{table}
	\centering
	\small
	\begin{tabular}{lc}
		\hline
		\bf Model & \bf Dev $F_1$   \\ 
		\hline
		\bf Best EE-GCN    & 67.17   \\
		\quad-- TDL    & 66.67   \\
		\quad-- NAEU     & 66.18  \\			
		\quad-- TDL \& NAEU        & 65.48   \\
		\quad--  MDER    & 66.40  \\
		\quad-- BiLSTM     & 63.87   \\
		\hline
	\end{tabular}
	\caption{ An ablation study of EE-GCN. TDL is short for typed dependency label, NAEU is short for node-aware edge update module, MDER is short for multi-dimensional edge representation. Scores are median of 5 models.}
	\label{tab:ablation}
\end{table}

\begin{figure}[t!]
	\centering
	\includegraphics[width=0.85\columnwidth]{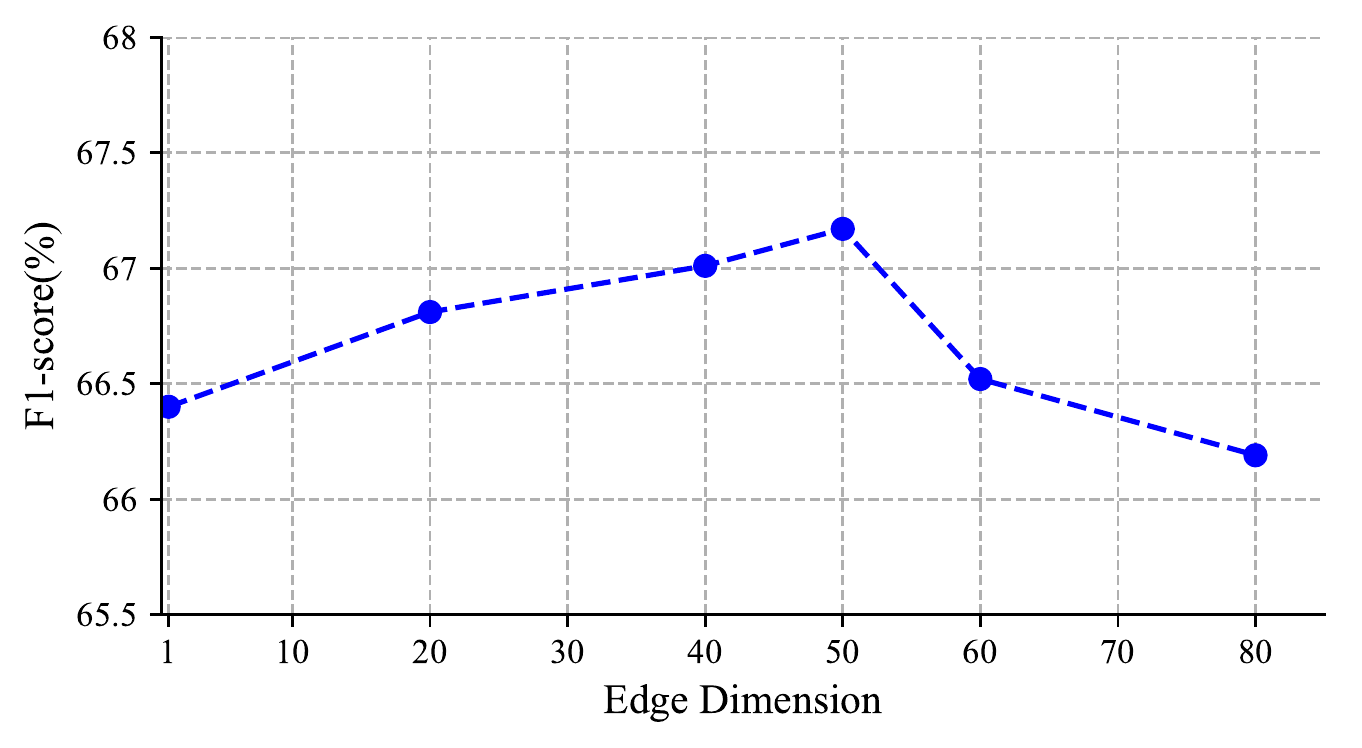}
	\caption{$F_1$-score variation with edge dim on dev.}
	\label{fig:dev-edge-dimension}
\end{figure}

\begin{figure}[!h]
	\centering
	\includegraphics[width=0.85\columnwidth]{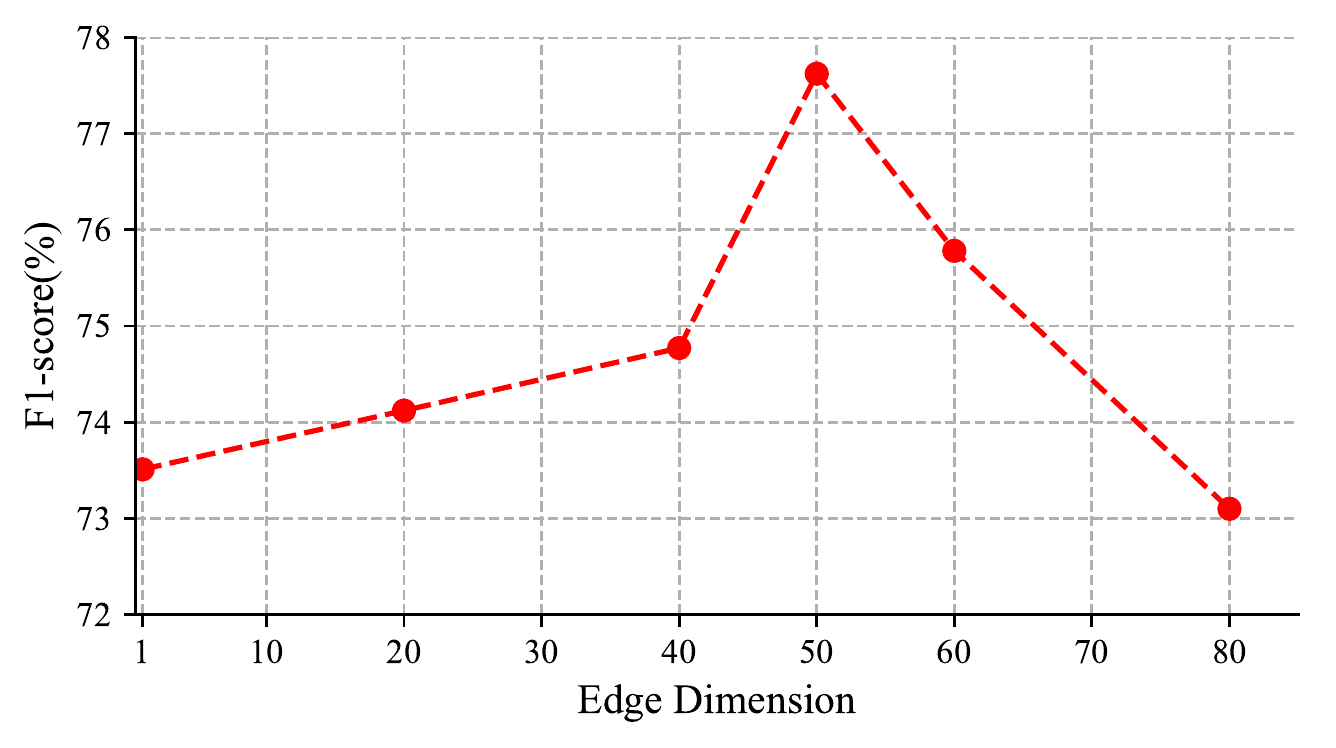}
	\caption{$F_1$-score variation with edge dim on test.}
	\label{fig:test-edge-dimension}
\end{figure}

\subsection{Effect of Edge Representation Dimension}

As shown in the ablation study, reducing the dimension of edge representation to 1 hurts the performance of EE-GCN deeply.
One may wonder what is the appropriate dimension for EE-GCN.
Therefore, we study the performance of the models with different dimensions of edge representation in this part. 
We vary the value of dimension from 1 to 80 with interval of 20 and check the corresponding $F_1$-score of EE-GCN on the dev and test set of ACE2005. 
The results on the ACE2005 dev and test set are illustrated in Figure~\ref{fig:dev-edge-dimension} and Figure~\ref{fig:test-edge-dimension} respectively.
We could see that the $F_1$-score peaks when the dimension is 50 and then falls.
This again justifies the effectiveness of introducing multi-dimensional edge representation.
Besides, the problem of overfitting takes effect when the dimension rises beyond a threshold, explaining the curve falls after the 50-dimensional representation in Figure~\ref{fig:dev-edge-dimension}.
%

\subsection{Effectiveness of Dependency Label}
To further confirm the effectiveness of dependency label, we add another experiment by adding dependency label to EEGCN-TDL individually. Based on F1=75.51\% on test set with removed TDL, the maximum improvements are F1=77.09\%, 77.22\% and 76.69\% when we respectively add dependency label of nmod, nsubj and dobj. This shows that these three labels are the mainly contributional labels, which is in consistent with our statistical in Introduction. 

\subsection{Performance of Different Event Types}
We reviewed F1-score of each type of events using EE-GCN and GCN respectively, and observe that End-ORG(F1=0.0) and Start-ORG(F1=41.67\%) are the hardest event types to detect for GCN. These two types of events gets significant improvement when using EE-GCN(F1=75.00\% for END-ORG and F1=71.43\% for Start-ORG), this demonstrates that the introducing dependency labels does help to improve ED. Besides, we notice that EE-GCN poorly performs on event types of ACQUIT, EXTRADITE and NOMINATE, which may be attributed to the very small amount of annotated instances of these types(only 6,7,12 respectively).
\subsection{Impact of EE-GCN layers}
\begin{figure}[t!]
	\centering
	\includegraphics[width=0.85\columnwidth]{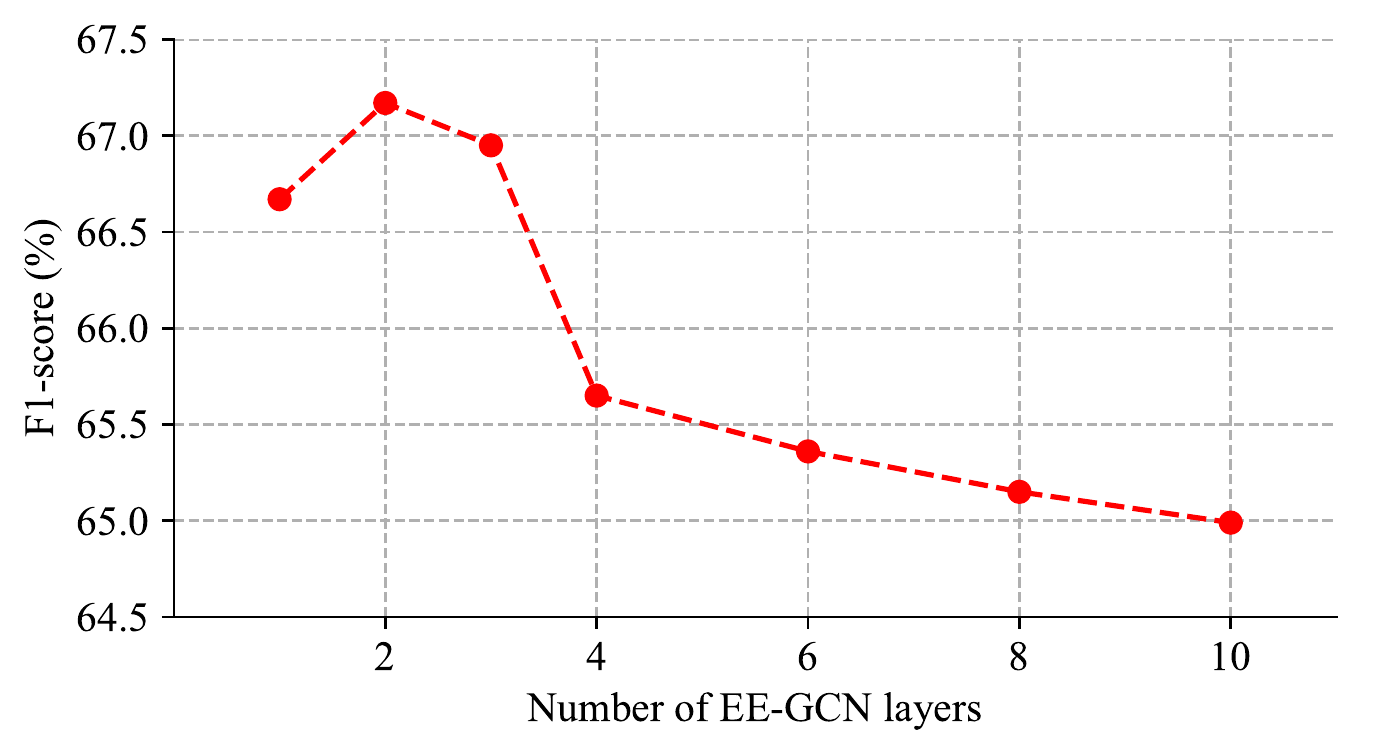}		
	\caption{$F_1$-score variation with GCN  layers on dev.}
	\label{fig:dev-layer}
\end{figure}

\begin{figure}[!h]
	\centering
	\includegraphics[width=0.85\columnwidth]{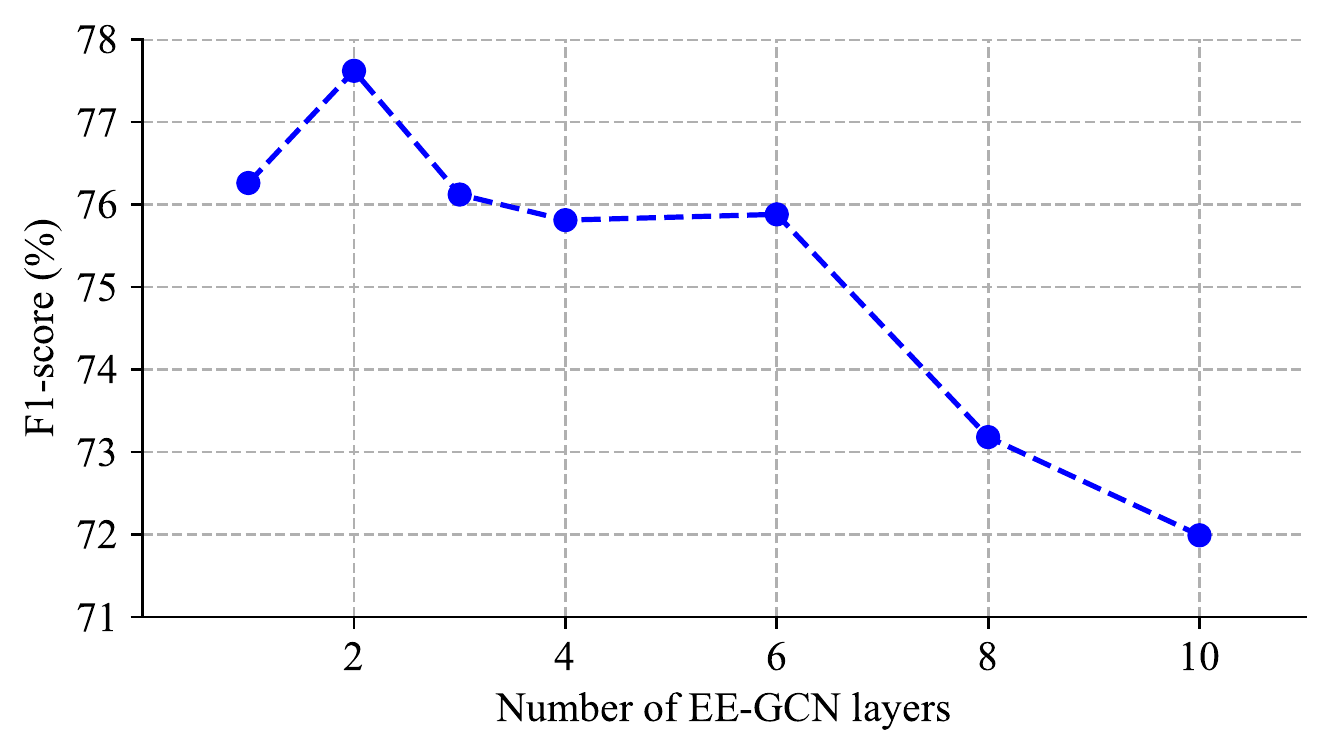}		
	\caption{$F_1$-score variation with GCN  layers on test.}
	\label{fig:layer}
\end{figure}

As EE-GCN can be stacked over $L$ layers, we investigate the effect of the layer number $L$ on the final performance.
Different number of layers ranging from 1 to 10 are considered. 
As shown in Figure~\ref{fig:dev-layer}, it can be noted that the performance increases with increasing EE-GCN layers. 
However, we find out EE-GCN encounters a performance degradation after a number of layers and the model obtains the best performance when $L=2$, so is the performance on test set in Figure~\ref{fig:layer}.
For this observation, two aspects are considered:
First, EE-GCN can only utilize first-order syntactic relations over dependency tree when $L=1$, which is not enough to bring important context words that are multi-hops away on the dependency tree from the event trigger into the trigger representation.
Second, EE-GCN operating on shallow dependency trees tends to over-smooth node representations, making node representations indistinguishable, thus hurting the model performance~\cite{zhou2018graph}.
%



\begin{figure*}[htb]
	\centering
	\subfigure[]{
		\begin{minipage}[h]{0.5\linewidth}
			\centering
			\includegraphics[scale=0.66]{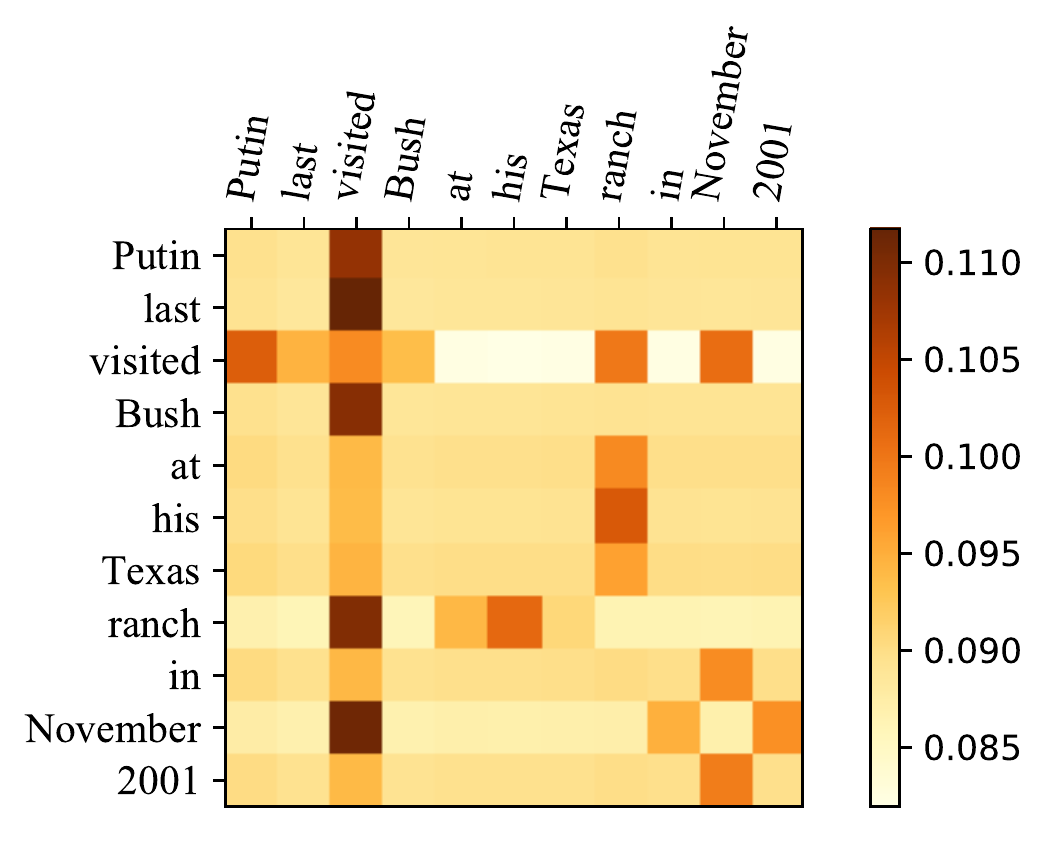}
		\end{minipage}%
	}%
	\subfigure[]{
		\begin{minipage}[h]{0.5\linewidth}
			\centering
			\includegraphics[scale=0.66]{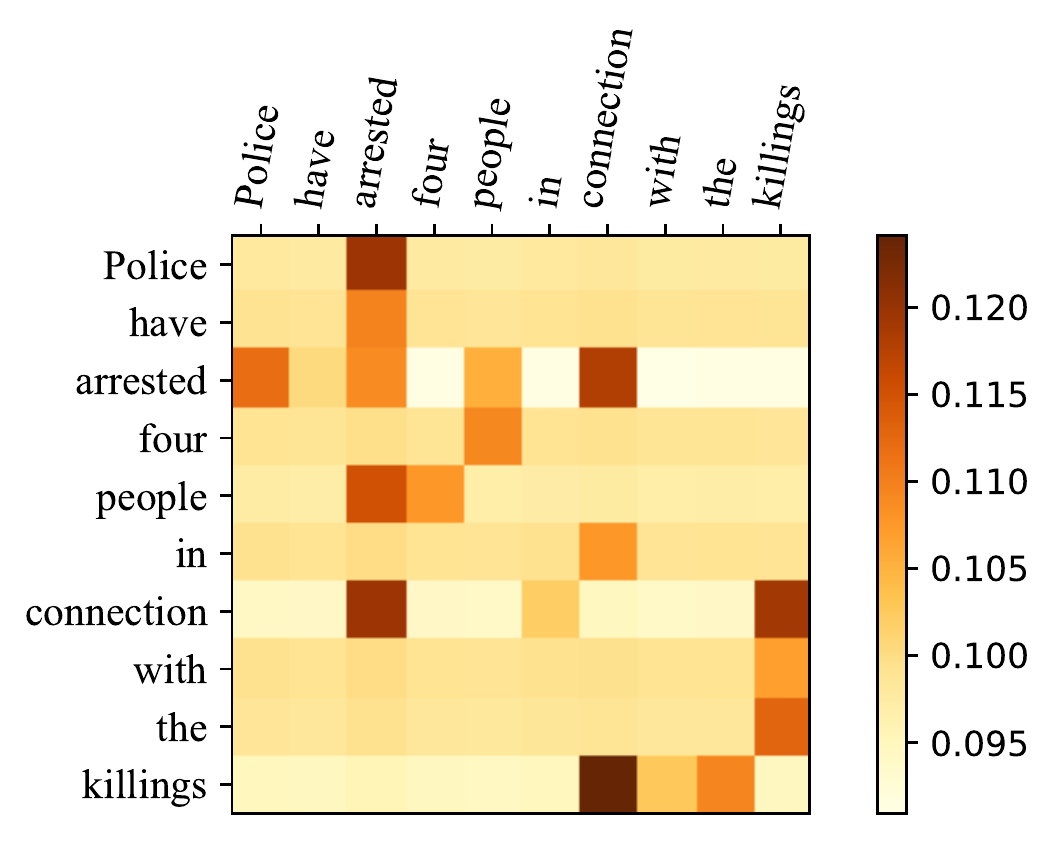}
		\end{minipage}%
	}%
	\caption{Visualization of the adjacency tensor $\mathbf{E}$ in EE-GCN.}
	\label{fig:case}
\end{figure*}

\subsection{Efficiency Advantage}

Since EE-GCN and RGCN both exploit syntactic structure and typed dependency labels simultaneously, we compare the efficiency of these two architectures from two aspects: parameter numbers and running speed.
For the sake of fairness, we run them on the same GPU server with the same batch size.
According to our statistics, the amount of parameters of EE-GCN and RGCN event detection architecture are 2.39M and 4.12M respectively.
Besides, EE-GCN performs 9.46 times faster than RGCN at inference time.
With the performance shown in Table~\ref{tab:performance}, we can conclude that EE-GCN not only achieves better performance, but also outperforms RGCN in efficiency.
This is mainly because EE-GCN exploits typed dependency labels by mapping them to relation embedding, while RGCN encodes different types of dependency labels with different convolutional filters.
Mathematically, given a graph with $r$ types of relations, the number of relation-related parameters in EE-GCN is only $p \times r$ while that in RGCN is $r \times h \times h$, where $p$ is the dimension of relation embedding and $h$ is the hidden state size of GCN.
Considering that $p$ and $h$ are usually set in the same order, the number of parameters in RGCN increases more rapidly than EE-GCN because $h \times h$ is significantly greater than $p$. 
We could also read from Table~\ref{tab:speed} that EE-GCN works almost as fast as GCN in both training and inference phrase, while RGCN works in a much slower way, which demonstrates that EE-GCN incorporated typed dependency relation without hurting efficiency badly. 


\begin{table}
	\centering
	\small
	\begin{tabular}{crrr}
		\hline
		\bf  & \bf GCN & \bf RGCN & \bf EE-GCN   \\ 
		\hline
		Training    & 15.7Bat/s   & 0.9Bat/s & 13.7Bat/s \\
		Inference   & 178.2Bat/s   & 17.7Bat/s   & 167.6Bat/s\\
		\hline
	\end{tabular}
	\caption{ Comparison of training/inference speed between GCN, RGCN and EE-GCN. Bat/s refers to the number of batches that can be processed per second.}
	\label{tab:speed}
\end{table}

\subsection{Case Study}

In this section, we present a visualization of the behavior of EE-GCN on two instances chosen from the ACE2005 test set, with the aim to validate our motivation provided in the introduction section.
We wish to examine whether EE-GCN indeed focuses on modeling the relationship between event-related words through a per instance inspection, which is shown in Figure~\ref{fig:case}.
Following~\cite{sabour2017dynamic}, we use the $l_2$ norm  of relation representation in the adjacency tensor of the last EE-GCN layer ($L=2$) to represent the relevance score of the corresponding word pair.
In the first case, each word has a high relevance score with ``\emph{visited}'' (the third column), because it is the event trigger.
This trigger has the strongest connections with ``\emph{Putin}'', ``\emph{ranch}'', ``\emph{November}'' and ``\emph{Bush}'' (the third row), which means that these four words are the top contributors for the detection of ``\emph{visited}'' in EE-GCN.
Similarly, in the second case, EE-GCN is able to precisely connect the event trigger ``\emph{arrested}'' with its subject ``Police'' and object ``\emph{people}''.
In general, the visualization result accords with the human behavior and shows the power of EE-GCN in capturing event-related relations between words.

\section{Conclusion and Future Works}

In this paper, we propose a novel model named Edge-Enhanced Graph Convolutional Networks (EE-GCN) for event detection.
EE-GCN introduces the typed dependency label information into the graph modeling process, and learns to update the relation representations in a context-dependent manner.
Experiments show that our model achieves the start-of-the-art results on the ACE2005 dataset.
%
%
In the future, we would like to apply EE-GCN to other information extraction tasks, such as relation extraction and aspect extraction.

\section{Acknowledgements}
We would like to thank the anonymous reviewers for their insightful comments and suggestions. 
This research is supported by the National Key Research and Development Program of China (grant No.2016YFB0801003) and the Strategic Priority Research Program of Chinese Academy of Sciences (grant No.XDC02040400).

\bibliography{anthology,emnlp2020}
\bibliographystyle{acl_natbib}

\end{document}